%% file: root.tex
\documentclass[letterpaper, 10 pt, conference]{ieeeconf}  

\IEEEoverridecommandlockouts                              

\usepackage{graphics} 
\usepackage{algorithm}  
\usepackage{algorithmic}
\usepackage{epsfig} 
\usepackage{mathptmx} 
\usepackage{times} 
\usepackage{amsmath} 
\usepackage{amssymb}  
\usepackage{subfigure}
\usepackage{diagbox}
\usepackage{threeparttable}
\usepackage{xcolor}

\newenvironment{myitemize}{\begin{list}{$\bullet$}
{\setlength{\topsep}{1mm}
\setlength{\itemsep}{0.25mm}
\setlength{\parsep}{0.25mm}
\setlength{\itemindent}{0mm}
\setlength{\partopsep}{0mm}
\setlength{\labelwidth}{15mm}
\setlength{\leftmargin}{4mm}}}{\end{list}}

\title{\LARGE \bf
TAE: A Semi-supervised Controllable Behavior-aware

Trajectory Generator and Predictor 
}

\author{Ruochen Jiao$^{1}$, Xiangguo Liu$^{1}$, Bowen Zheng$^{2}$, Dave Liang$^{2}$, and Qi Zhu$^{1}$
\thanks{$^{1}$Ruochen Jiao, Xiangguo Liu, and Qi Zhu are with the Department of Electrical and Computer Engineering, Northwestern University, Evanston, IL, 60201, USA.
    {\tt\small 
       \{ruochen.jiao, xg.liu\}@u.northwestern.edu, 
       qzhu@northwestern.edu.}}%
\thanks{$^{2}$Bowen Zheng and Dave Liang are with Pony.ai, Fremont, CA 94538, USA.
        {\tt\small \{bowen.zheng, dave.liang\}@pony.ai.}
        }
\thanks{We gratefully acknowledge the support from NSF grants 1834701, 1839511, 1724341, 2038853, and ONR grant N00014-19-1-2496.}
}

\begin{document}

\maketitle
\thispagestyle{empty}
\pagestyle{empty}

\begin{abstract}
Trajectory generation and prediction are two interwoven tasks that play important roles in planner evaluation and decision making for intelligent vehicles. Most existing methods focus on one of the two and are optimized to directly output the final generated/predicted trajectories, which only contain limited information for critical scenario augmentation and safe planning. In this work, we propose a novel behavior-aware Trajectory Autoencoder (TAE) that explicitly models drivers' behavior such as aggressiveness and intention in the latent space, using semi-supervised adversarial autoencoder and domain knowledge in transportation. Our model addresses trajectory generation and prediction in a unified architecture and benefits both tasks: the model can generate diverse, controllable and realistic trajectories to enhance planner optimization in safety-critical and long-tailed scenarios, and it can provide prediction of critical behavior in addition to the final trajectories for decision making. Experimental results demonstrate that our method achieves promising performance on both trajectory generation and prediction.

\end{abstract}

\section{Introduction}

Tremendous progress has been made for enabling autonomous driving in recent years. The autonomous driving pipeline typically consists of several modules such as sensing, perception, prediction~\cite{liu2021multimodal,ye2021tpcn}, planning~\cite{liu2020trajectory,liu2022markov,liu2022neural}, and control, which can be roughly divided as two parts -- environment perception and decision making. Between these two parts, the prediction and generation of surrounding vehicles' trajectories can be viewed as a two-way bridge. 
In the forward direction, the prediction module encodes the environment information and translates it into potential future trajectories of surrounding vehicles to facilitate the planning module. Reversely, to train and evaluate the planning module, we will need to discover critical traffic scenarios and vehicle behaviors, and generate more realistic and diverse trajectories of surrounding vehicles -- this is particularly important for evaluating the safety of vehicle planning as some of the ``long-tail'' scenarios could be quite challenging and lead to the violation of safety requirements~\cite{zhu2021safety,zhu2020know,jiao2021end}.

Most existing works of trajectory generation or augmentation indeed try to identify risky scenarios and then extract corresponding features or styles in order to generate more safety-critical scenarios. For instance, in~\cite{ding2020cmts} and~\cite{yin2021diverse}, the authors extract features and variables that lead to safety-critical scenarios and then feed them into generative models. 
However, the definition of the critical styles is vague, and the controllability and interpretability of the models are limited.  \cite{yin2021diverse} also points out that most existing works focus on generating the entire scenarios but lack control over individual agents (vehicles) and their detailed behaviors. 

\begin{figure}[!ht]
    \centering
    \includegraphics[width=\columnwidth]{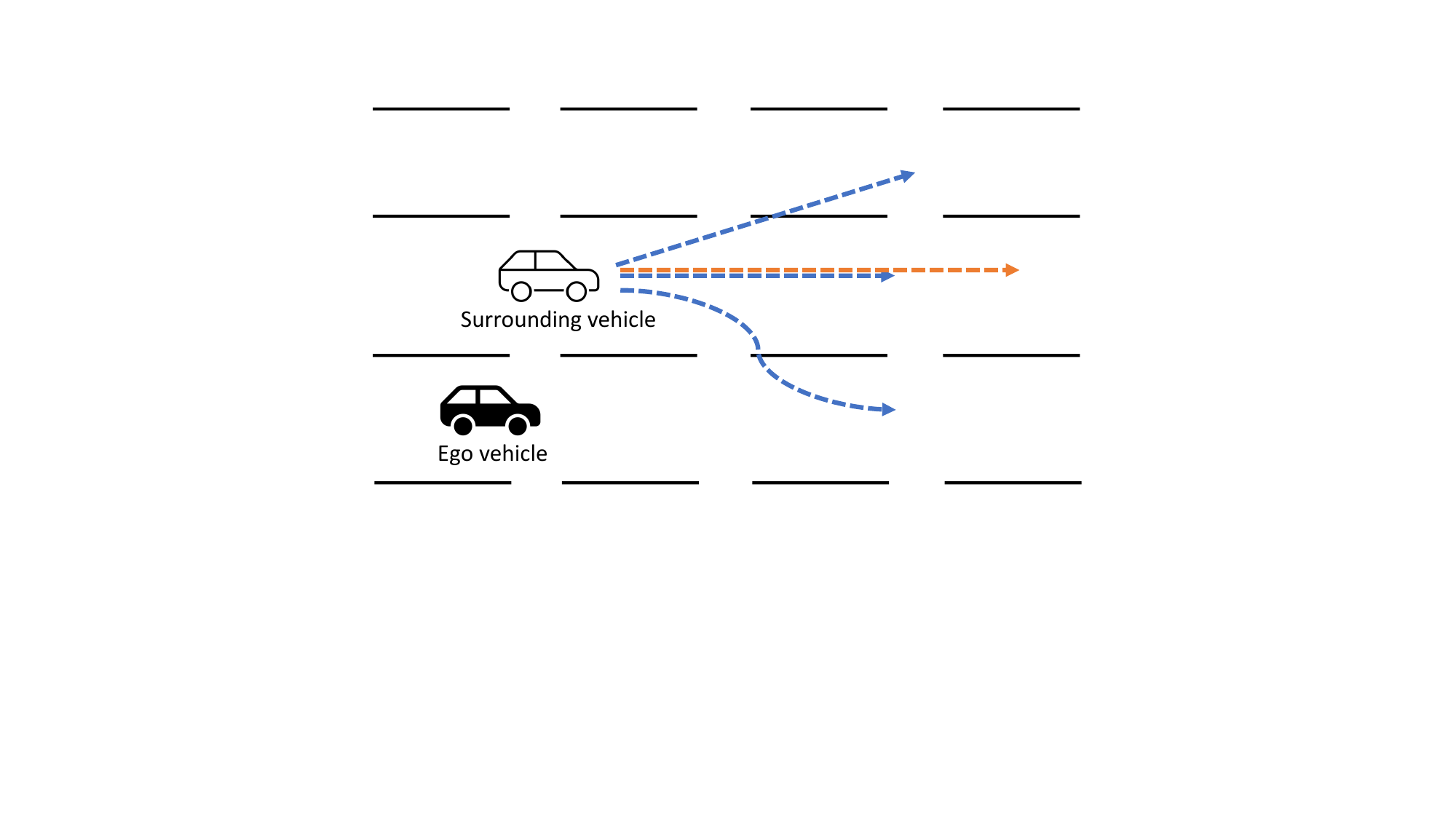}
    \caption{Trajectories with different intentions and aggressiveness levels. Blue dashed lines demonstrate different potential intentions in changing lanes and the orange one shows a more aggressive trajectory in the current lane.}
    \label{fig:semi-aae}
\end{figure}

On the prediction side, recent works on motion forecasting~\cite{liu2021multimodal, liang2020learning, zhao2020tnt, ngiam2021scene,ye2021tpcn,gilles2021gohome}  have been focusing on the displacement error between the predicted trajectories and the ground truth, with great performance achieved. However, as mentioned in~\cite{wilson2021argoverse}, the performance on current datasets has begun to plateau. Moreover, other than the trajectories themselves, those works pay little attention to other information that could also be very important for understanding surrounding vehicle's behaviors and making safe decisions. For instance, behavior prediction such as whether other vehicles may change lanes or whether they may yield to the ego vehicle, is critical information for safe planning~\cite{liu2022physics}. In fact, human drivers make decisions generally relying on high-level predictions instead of exact future trajectories of surrounding vehicles. 
Thus, in our work, we consider \emph{intention and aggressiveness} as such high-level behaviors of surrounding vehicles -- their detailed definitions are explained in Section~\ref{sec: methods} but Fig.~\ref{fig:semi-aae} shows a simple illustration -- and leverage them in both predicting trajectories and in generating more diverse trajectories and behaviors. To the best of our knowledge, this is the first work that explicitly models and utilizes aggressiveness in trajectory prediction and generation.

Unlike previous methods that are designed solely for either trajectory prediction or generation, we propose a unified framework for both tasks using behavior-aware adversarial autoencoder architecture combined with domain knowledge in the transportation. Our goal is to design a \emph{hierarchical and behavior-aware} predictor and a generator that can augment \emph{realistic, diverse, explainable, and controllable} trajectories. We believe that this will facilitate both prediction and planning modules to address the critical (and potentially unsafe) tail events on the road. More specifically, our contribution can be summarized as follows:
\begin{myitemize}
\item We propose Trajectory Autoencoder (TAE), a novel and unified architecture based on adversarial autoencoder for trajectory generation and prediction. It facilitates both tasks with behavior-level awareness and control. 

\item Ours is the first work to explicitly consider aggressiveness in trajectory generation/prediction. We utilize semi-supervised training along with adversarial generation and domain knowledge to model the behaviors with limited data. The method is extensible for other driving behaviors.

\item We conduct experiments in a commonly-used dataset for trajectory generation and prediction. We evaluate five metrics to demonstrate the advantages of our methods in generating diverse and controllable vehicle trajectories and safety-critical scenarios, and in predicting surrounding vehicles' behaviors.

\end{myitemize}

The  rest  of  the  paper  is  organized  as  follows. Section~\ref{section:background} reviews the related works. Section~\ref{sec: methods} presents the methodology and major components of our proposed semi-supervised behavior-aware TAE. Section~\ref{sec: experiments} presents the experimental results and discussions. Section~\ref{sec:conclusion} concludes the paper.


\input{background}

\input{method}
\input{experiments}

\section{Conclusion}
\label{sec:conclusion}
In this work, we propose a behavior-aware trajectory autoencoder (TAE) for both vehicle trajectory generation and prediction. We embed the domain knowledge such as intention and aggressiveness into the latent space and optimize the model with limited labelled data. Our method can generate realistic, diverse, and controllable trajectories, which could greatly benefit reliable decision making and planning evaluation in critical scenarios.

\bibliographystyle{ieeetran}
\bibliography{reference}

\end{document}

%% file: background.tex
\section{Background}
\label{section:background}
\subsection{Trajectory Generation and Prediction}
\subsubsection{Trajectory Generation}
Trajectory generation or augmentation is of great significance to optimize and evaluate decision making module in autonomous driving. \cite{ding2021multimodal} proposes a flow-based generative model
using the objective function of weighted likelihood to generate multimodal safety-critical scenarios. Their following work \cite{ding2020cmts} demonstrates a generative model conditioned on road maps to bridge safe
and collision driving data. The model combines conditional variational autoencoder and style transferring techniques to generate the whole risky scenario, but it cannot control agent-level trajectories. \cite{yin2021diverse} proposes a RouteGAN to generate diverse trajectories for every single agent and the trajectory is influenced by a style variable. However, the latent spaces of these generative methods are not well explained with driving or transportation knowledge, especially at the behavior level. And because of the nature of GAN and style transferring techniques, the models only have rough and limited control over the generation process, which may lead to unrealistic and uncontrolled trajectories.  
\subsubsection{Trajectory Prediction}
Recent works have applied different methods to represent the past trajectory and contexts. CNN with rasterized images \cite{phan2020covernet},  graph neural networks (GNN) \cite{liang2020learning, zhao2020tnt},  transformers \cite{yuan2021agentformer, liu2021multimodal}, and even 3D point cloud \cite{ye2021tpcn} are used to encode the map and interaction information. These works achieve good performance in terms of the displacement error between ground truth and prediction. In our work, we choose to use GNN-based method for extracting features, but with the additional consideration of behavior-level prediction for safety-critical scenarios. 


\subsection{Driving Behavior Modeling}
The work in~\cite{8500419} proposes an intention predictor based on mixture density network (MDN)~\cite{bishop1994mixture}, which considers the semantic information on the road and predicts the insertion area (region proposals) using the MDN architecture. However, the approach has to select useful features and define the insertion area manually in different scenarios. \cite{ding2019online} designs an online two-level framework that anticipates the high-level driving policy such as forward, yielding, turning, and then feeds such intention to an optimization-based predictor. 

Besides the intention of changing lane and turning, the vehicle's style such as aggressiveness will also influence its motion. Generally, aggressive vehicles tend to drive at higher acceleration. Most works measure aggressiveness based on sensors'(e.g., accelerator, gyroscope) data \cite{8778416,ma2018comparative} or long-term statistics~\cite{gonzalez2014modeling}. Some works propose online measurement in different scenarios. \cite{yu2018human} demonstrates an aggressiveness measurement in lane changing scenario by using a combination of space utility and safety utility, where the space utility is the available space for lane changing while the safety utility is measured by the time headway of a vehicle, which can be generalized to other driving scenarios.

\subsection{Adversarial Autoencoder}

\begin{figure}[!ht]
    \centering
    \includegraphics[width=1\columnwidth]{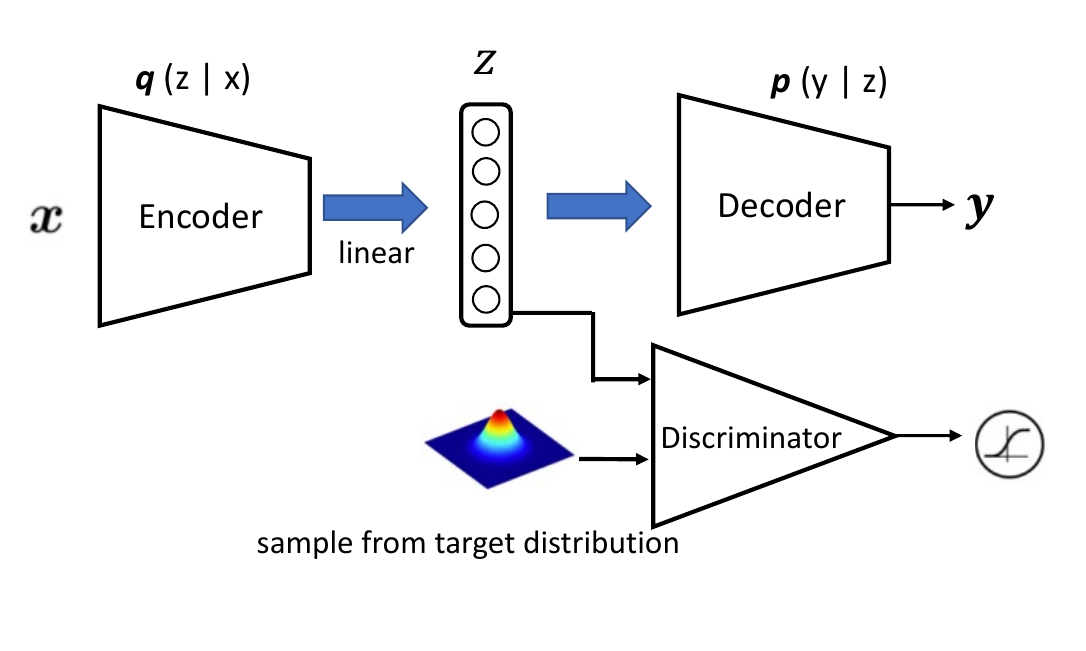}
    \caption{Architecture of the basic adversarial autoencoder (AAE).}
    \label{fig:base-aae}
\end{figure}

\begin{figure*}[th]
    \centering
    \includegraphics[width=1.8\columnwidth]{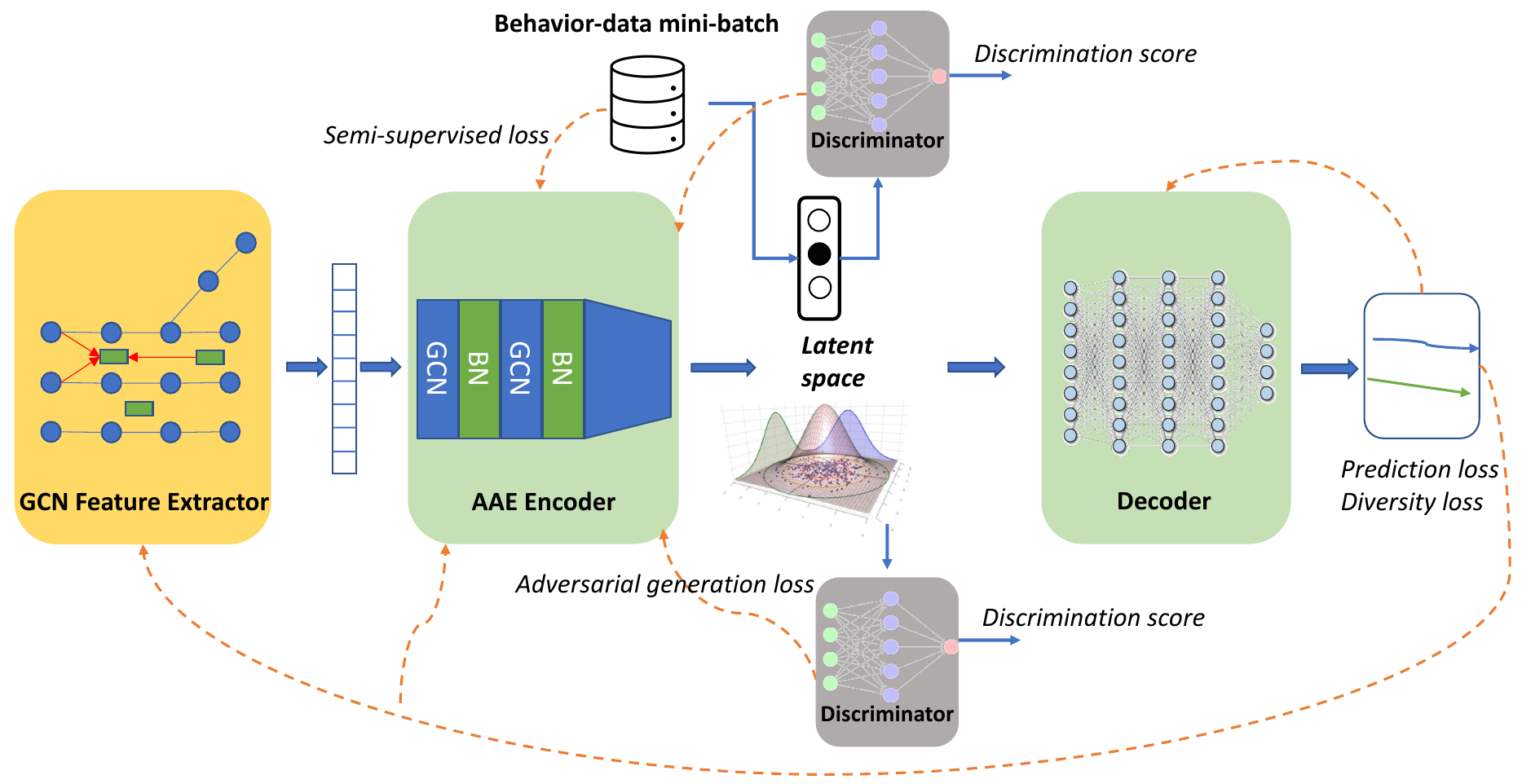}
    \caption{Overview of our proposed TAE architecture and training pipeline.}
    \label{fig:pipeline}
\end{figure*}

The variational autoencoder~(VAE)~\cite{kingma2013auto} provides a principled method for jointly learning deep latent-variable models and corresponding inference models using stochastic gradient descent~\cite{kingma2019introduction}. Training a VAE model consists of two steps: regularization and reconstruction. The regularization step is aimed to encode the input as certain distributions (usually Gaussian) over the latent space using Kullback-Leibler (KL) divergence, while the reconstruction step is used to decode the latent variables to the target space. 
In contrast to VAE that uses KL divergence and evidence lower bound, adversarial autoencoder (AAE)~\cite{makhzani2015adversarial} uses adversarial learning for imposing a specific distribution on the latent variables. 
The AAE architecture is superior to VAE in terms of imposing complicated distributions and shaping the latent space.

More specifically, let $x$ be the input, $y$ be the output, and $z$ be the latent vector of an autoencoder with a deep
encoder and decoder. Let $p(z)$, $q(z|x)$, $p(y|z)$ and $p_d(x)$ denote the prior distribution on the latent vectors, encoding distribution, decoding distribution and input data distribution, respectively. 
The encoding function of the autoencoder $q(z|x)$ defines an
aggregated posterior distribution of $q(z)$ on the latent vector of the autoencoder as follows:
\begin{equation}
q(\mathbf{z})=\int_{\mathbf{x}} q(\mathbf{z} \mid \mathbf{x}) p_{d}(\mathbf{x}) d \mathbf{x}\label{vae_eq}
\end{equation}

The $q(z)$ is expected to match the prior distribution $p(z)$. In the AAE, the encoder tries to fool the discriminators into thinking the generated latent vectors are from the prior target distribution $p(z)$.

As shown in Fig.~\ref{fig:base-aae}, the AAE architecture maps the input $x$ to the latent space of a Gaussian distribution. The real data sampled from the Gaussian distribution and the latent codes are fed into the discriminator. The discriminator tries to distinguish the real samples from the generated ones, and the discrimination scores are used to update the encoder to generate data following target distribution. Then the decoder reconstructs the output from the latent code $z$. In our work, we will extend the AAE architecture to model multiple and complex distributions, and encode label information in the latent space.


%% file: method.tex
\section{Our Proposed TAE Architecture}
\label{sec: methods}

 The design of our proposed TAE architecture for trajectory generation/prediction is shown in Fig.~\ref{fig:pipeline}. 
  In this section, we will explain the major modules in our framework and the methodology for modeling and optimization, i.e., the context feature extractor (\ref{context}), the architecture of the semi-supervised behavior-aware AAE (\ref{aae}), the latent space modeling with prior knowledge of vehicle's behavior (\ref{latent}), the optimization pipeline (\ref{objective}), and several additional improvements (\ref{diverse}).

\subsection{Context Modeling}
\label{context}

\begin{figure}[!ht]
    \centering
    \includegraphics[width=1\columnwidth]{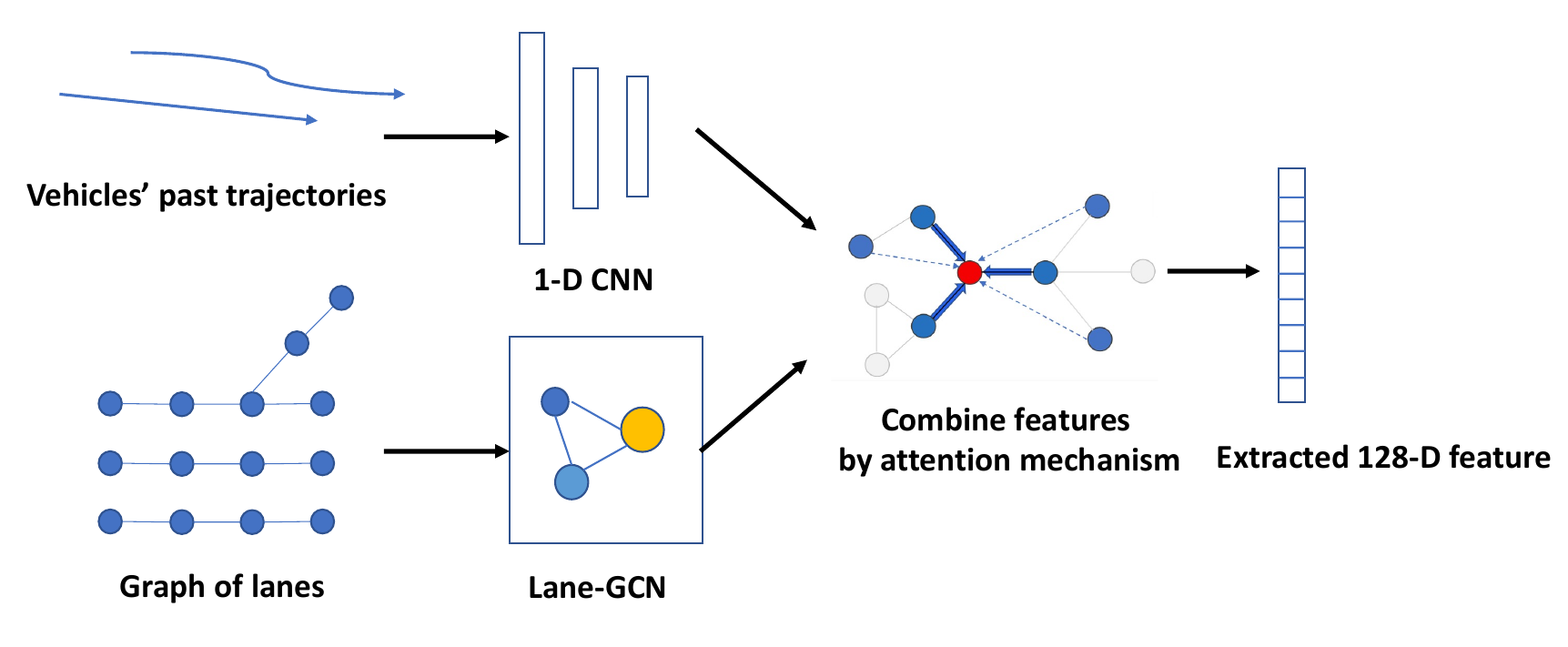}
    \caption{Feature Extractor: extract trajectories using 1-D CNN; model interactions and structured map information by GCN \cite{liang2020learning}.}
    \label{fig:extractor}
\end{figure}

In order to capture features of the environment, we need to consider the past trajectories, the interactions between vehicles, and the map information. Similar to \cite{liang2020learning}, we use the one-dimensional dilated convolutional neural network to extract features from history trajectories and utilize the graph convolutional network (GCN) to model the graphed map information and the interaction. The GCN-based method has shown good performance in modeling transportation contexts since most vehicles drive on the structured roads, especially in urban scenarios. Then, the feature extractor applies an attention mechanism to combine the information and outputs a 128-dimensional feature for each agent. The pipeline of the feature extraction is shown in Fig.~\ref{fig:extractor}.

 \subsection{Behavior-aware Semi-supervised AAE}
\begin{figure}[!ht]
    \centering
    \includegraphics[width=1\columnwidth]{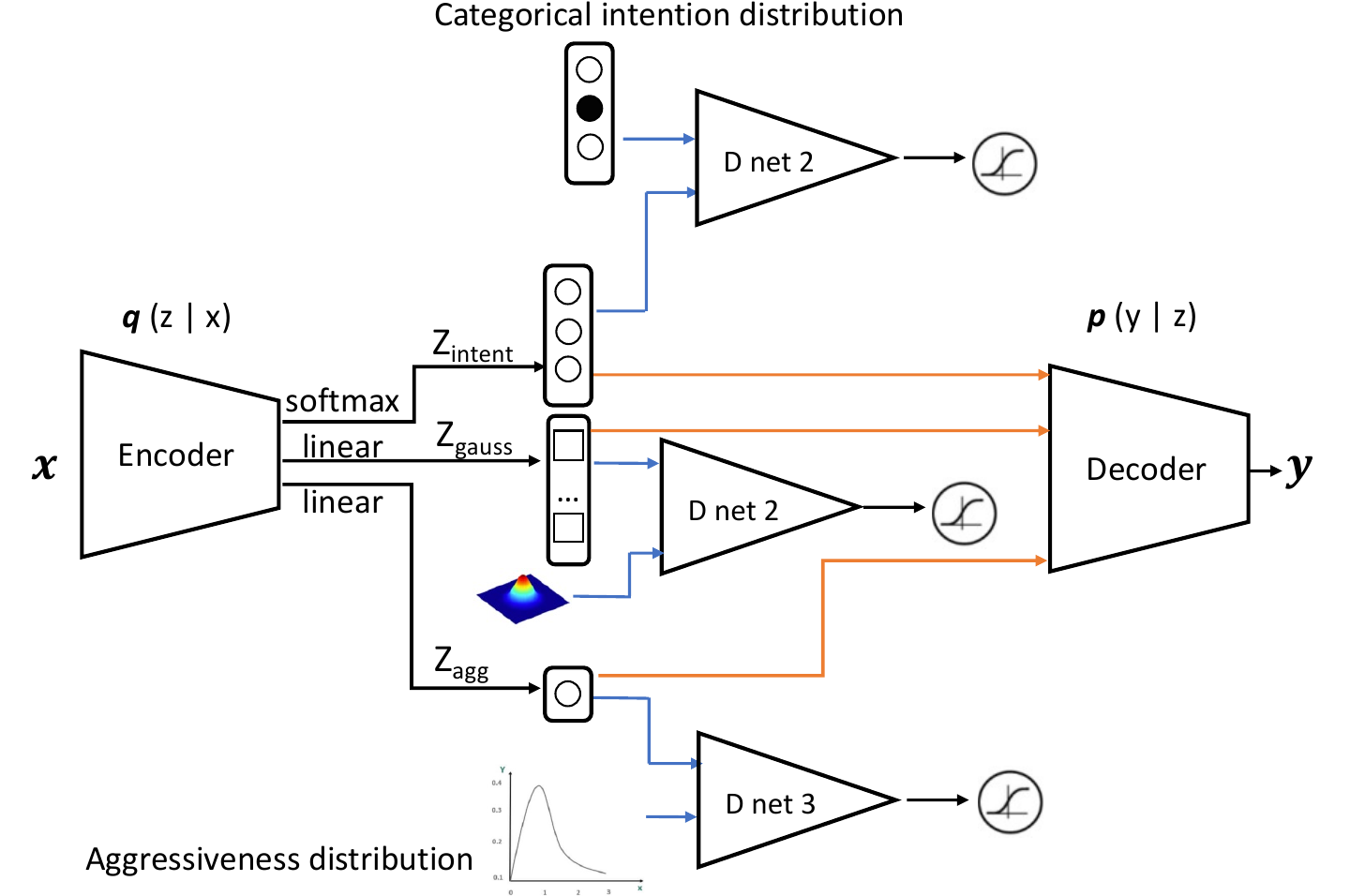} 
    \caption{Behavior-aware adversarial autoencoder.}
    \label{fig:proposed_aae}
\end{figure}

 The AAE architecture itself blends the autoencoder architecture with the adversarial loss concept introduced by GAN, and replaces the KL divergence in VAE with adversarial loss to regularize diverse and complex distributions of latent space. In our model, we further utilize semi-supervised learning to model the driving behavior in the latent space by incorporating the limited label information.  The architecture of the proposed semi-supervised behavior-aware AAE is shown in Fig.~\ref{fig:proposed_aae}. 
 The model consists of an encoder, behavior-aware and remaining latent space, discriminators for different latent vectors, and a decoder.
\label{aae}
 
The input $x$ is a 128-dimensional feature generated by the GCN-based feature extractor. The multi-head encoder based on a two-layer GCN projects the features to a lower-dimensional latent space. 
 
Our proposed AAE model has three parts of the latent space, which follow different distributions. They are three-dimensional intention latent vector $z_{intent}$, following categorical distributions, one-dimensional aggressiveness latent vector $z_{agg}$, following a log-normal distribution, and remaining latent vector $z_{gauss}$, following Gaussian distributions. For the dimension of the remaining latent vector, we notice a trade-off between the generated trajectory's smoothness and behavior's controllability and we allocate six dimensions to the remaining latent space after preliminary experiments.

For each group of latent variables, we input the samples from real target distribution and the generated latent variables to the corresponding discriminators (D net in Fig.~\ref{fig:proposed_aae}) to regularize the latent distribution. For instance, the discriminator for intention latent vector is trained to distinguish our generated latent vector from the sample in real categorical distribution. By adversarial learning, we can force the three latent vectors to follow their corresponding distributions.

After the adversarial generation learning stage, we collect data and labels for the semi-supervised aggressiveness and intention modeling. Only part of the vehicles' behaviors can be identified and labeled, but in general, these behaviors are in certain distributions. This is the reason why the semi-supervised learning works for training the behavior vectors. In the semi-supervised training mini-batch, we optimize the encoder to predict the real intentions and aggressiveness levels in the latent space, based on the limited labelled data. 

Besides regularizing the latent space and optimizing the encoder, the model needs to generate realistic trajectories. The latent vectors are concatenated and fed to the decoder. The decoder is a three-layer fully-connected network that maps the latent vectors into future trajectories. Finally, we update the whole pipeline including feature extractor, encoder and decoder to make the generated trajectory close to the reference. 

The details of behavior modeling and multi-stage optimization process are introduced in the following sections.

\subsection{Latent Space Modeling}
\label{latent}
Most previous works directly predict or generate the waypoints for future trajectories and the VAE-based works generally use unified latent variables.

With our semi-supervised AAE architecture, we can represent both distribution and label information of important driving features in the latent space if we define them in a learnable way. 
In this work, we have two behavior latent vectors, aggressiveness and intention. 
\subsubsection{Aggressiveness}
Aggressiveness is an important feature of vehicle's behavior.  Conservative vehicles and aggressive vehicles may take different actions even in the identical scenario. However, it is still an open question to measure and predict the aggressiveness, especially in a general setting. As mentioned in Section~\ref{section:background}, some recent works~\cite{yu2018human, liu2020impact} propose different measurements of aggressiveness in specific scenarios such as lane changing or merging. In our work, we consider time headway as a common measurement when building the aggressiveness model, which measures the time difference between two successive vehicles when they cross a given point. 
We believe that time headway is a feature that we can capture in most driving scenarios and it can stand for vehicles' aggressiveness, especially in the longitudinal direction. Intuitively, the shorter the time headway is, the more aggressive the driving behavior is. We learn such attribute in a semi-supervised way because only some vehicles have close and observable interaction with the vehicle in front of them but we can assume every vehicle has its own intrinsic level of aggressiveness that follows a general distribution.

\begin{figure}[!ht]
    \centering
    \includegraphics[width=0.9\columnwidth]{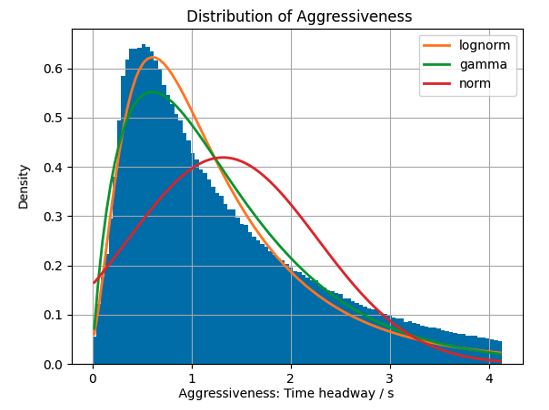}
    \caption{The histogram and fitted distribution of valid aggressiveness (time headway) in the Argoverse motion forecasting dataset.}
    \label{fig:agg-dist}
\end{figure}

To model the latent variable of aggressiveness by proposed AAE, we should have prior knowledge of the distribution of the aggressiveness, which could be influenced by many different factors such as traffic scenarios and vehicles' behaviors~\cite{ha2012time}. Log-normal distribution, Gamma distribution and normal distributions are potential distribution to model the time headway in various scenarios. Here we focus on the urban scenario and calculate time headway of all valid cases in the Argoverse~\cite{chang2019argoverse} motion forecasting dataset. The data histograms and fitted distributions are shown in Fig.~\ref{fig:agg-dist}.

We notice that the log-normal distribution fits the aggressiveness data best with the lowest KL divergence (0.017) and sum of squared error (0.16). In our model, we use the log-normal distribution as the prior distribution for the discriminator of aggressiveness latent vector.

In the semi-supervised learning phase, we collect labelled aggressiveness mini-batch and train the latent vectors to match their true values as a regression problem.

\subsubsection{Intention}
In our work, we represent intentions by three simple but reasonable classes: moving forward, turning/changing lane to the left, and turning/changing lane to the right. We are inspired by human-driving vehicles that inform other vehicles of the ego vehicle's intentions by using turn signals. These three intentions are discrete by nature and we model them with categorical distribution. We only label the vehicles that show clear intention in a long enough time frame (5 seconds in experiments).

\subsection{Optimization Pipeline}
\label{objective}
To produce realistic, diverse and 
controllable trajectories,  our model is designed to optimize and balance several different targets. The optimization process consists of three phases: prediction phase, regularization phase and semi-supervised phase. 

First, in the prediction phase, the whole model including the feature extractor, encoder and decoder is optimized to produce accurate and realistic trajectories. We apply the smooth L1 loss as shown below in~\eqref{l1-smooth} on all time steps to calculate the distance between the generated trajectory $\hat{y}$ and ground truth $y$.

\begin{equation}
Loss_{pred}\left(y_{i},\hat {y_i}\right)= \begin{cases}0.5 (y_{i}-\hat{y_i})^{2} & \text { if }\left\|y_{i}-\hat{y_{i}}\right\|<1 \\ \left\|y_{i}-\hat{y_i}\right\|-0.5 & \text { otherwise }\end{cases} \label{l1-smooth}
\end{equation}

To model the latent space, we apply both the adversarial learning loss and the semi-supervised learning loss. We utilize three different generators and discriminators to regularize the distribution of the latent space by adversarial learning. The adversarial regularization  loss is shown in~\eqref{eq:adv-loss}. Here $x$ represents the input of the encoder $G$, and $m$ equals 3, corresponding to the three distributions: Gaussian, Log-normal and Categorical distributions.




\begin{equation}
Loss_{adv}(x) = \frac{1}{m} \sum_{i=1}^{m} \log \left(1-D_i\left(G_i\left(x\right)\right)\right)\label{eq:adv-loss}
\end{equation}

For the discriminator $D$, we train them by maximizing the average of the log probability of real latent samples $s$ and the log of the inverse probability for fake latent samples:
\begin{equation}
Loss_D(x,s) = \log D_i\left(s_i\right)+\log \left(1-D_i\left(G_i\left(x\right)\right)\right)\label{eq:dis}
\end{equation}

In the semi-supervised phase, we update the encoder on labelled data of the behavior-aware latent space to make the latent variables explainable.  
To model the latent variable of aggressiveness $z_{agg}$, the encoder is trained to minimize the mean square error of the predicted aggressiveness and labeled ones $l_{agg}$. To represent the intention vector $z_{int}$, the encoder is updated to minimize the cross entropy cost on a labeled mini-batch (class labels are $l_{int}$), which is modeled as a classification problem. The total loss of semi-supervised learning phase is shown in equation~\eqref{eq:semi-loss}.

\begin{equation}
Loss_{Semi}(z_{agg},z_{int},l_{agg},l_{int}) =  -\sum_{i=1}^3l_{int}\log z_{int} + (l_{agg} - z_{agg})^2 \label{eq:semi-loss}
\end{equation}


The optimization pipeline is illustrated as Algorithm~\ref{alg:pipeline}.

\begin{algorithm}[htbp]
\caption{Optimization Pipelines}
\label{alg:pipeline}
\begin{algorithmic}[1]
\STATE\textbf{Initialize:} feature extractor $F$, AAE encoder $G$, decoder $R$, discriminator $D_{i}$, target distribution $p_{i}$, $i$ = 1,2,3.

\STATE\textbf{Input:} past trajectories $t$ and map graph $m$.

\FOR{each batch}
\STATE Let features $x$ = $F(t,m)$.
\STATE Let latent vectors $z$ = $G(x)$.
\STATE Sample $s_i$ from target distribution $p_{i}$ and calculate  $D_{i}(z_i)$ and $D_{i}(s_i)$.
\STATE Update $G$ by adversarial generation loss $Loss_{adv}$ \eqref{eq:adv-loss}.
\STATE Update $D_i$ by discrimination loss $Loss_D$ \eqref{eq:dis}.
\STATE Obtain the labelled mini-batches for intention and aggressiveness, respectively.
\STATE Calculate $Loss_{semi}$ \eqref{eq:semi-loss} and update $G$.
\STATE Concatenate the latent vectors and feed to decoder $R$ $\hat y$ = $R(z)$.
\STATE Calculate the prediction loss $Loss_{pred}(y,\hat y)$ and update $F$, $G$, $R$.

\ENDFOR

\end{algorithmic}
\end{algorithm}

\subsection{Diverse Generation and Multi-modal Prediction}
\label{diverse}
In our preliminary tests, we noticed that our model only had limited capacity to generate diverse trajectories, even though we had already shaped and trained the latent space to model the aggressiveness and intention using our proposed architecture.  We found that in the training, the sampling was aimed to maximize the likelihood that may only produce samples corresponding to the major modes of the data distribution~\cite{yuan2020dlow}. We also did not have control over all latent variables, and particularly, we only used one-dimensional variable to represent the aggressiveness. To address these problems, we introduce a diversity-promoting
prior over samples as a diversity objective to optimize the latent mappings for improving sample and decoding diversity. We calculate the diversity loss as in the equation~\eqref{eq:diversity_loss}~\cite{yuan2020dlow} based on a pairwise Euclidean distance among generated trajectories. In~\eqref{eq:diversity_loss}, $x_{i}$ is the $i$-th generated trajectory and $\sigma_{d}$ is used to normalize the distance.

\begin{equation}
Loss_{d}(X)=\frac{1}{K(K-1)} \sum_{i=1}^{K} \sum_{j \neq i}^{K} \exp \left(-\frac{D^{2}\left(\mathbf{x}_{i}, \mathbf{x}_{j}\right)}{\sigma_{d}}\right)\label{eq:diversity_loss}
\end{equation}

In the diversity optimization stage, we sample different behaviors and feed corresponding latent vectors to the decoder. By combining this loss with prediction loss, we can promote the generation diversity of different modes and improve the aggressiveness' control over the trajectories.

For trajectory prediction, we add an additional classifier to select a most possible trajectory from different ones, which enhances the performance of multi-modal prediction.

%% file: experiments.tex
\section{Experimental Results}
\label{sec: experiments}
\subsection{Experiment Settings}
We train our model on the Argoverse motion forecasting dataset \cite{chang2019argoverse} and evaluate the generation and prediction performance on the corresponding validation and test sets. The Argoverse motion forecasting benchmark has more than 30K scenarios collected in Miami and Pittsburgh. Each scenario has detailed graph of road map and multiple agent trajectories sampled at 10 Hz. In the motion forecasting and generation tasks, trajectories of the first 2 seconds are offered as input data. The dataset contains the straight road and intersection scenarios, most of which are easy and safe cases. 

We train our model on an NVIDIA Titan RTX platform for 30 epoches. The batch size is 32. The learning rates are set as $1e-4$, $1e-5$, $1e-5$ and $5e-5$ for the Adam optimizers of prediction, adversarial generation learning, discrimination and semi-supervised phases, respectively.

\subsection{Diverse and Controllable Trajectory Generation}
To measure the performance of the trajectory generation, we 1) calculate the cluster numbers of the dataset to evaluate the augmented complexity and diversity, 2) visualize the generated trajectories to demonstrate the controllability and interpretability in generation, and 3) sweep the behavior latent space and count the risky cases to test the capability of generating safety-critical scenarios.

 First, we cluster normalized generated trajectories and obtain the cluster numbers with different thresholds. Generally, a larger cluster number represents a higher level of complexity and diversity, while the threshold constrains the minimum proportion the clusters.  Since we do not have clear labels for generated trajectories, we utilize Dirichlet Process Gaussian Mixture Models (DPGMM)~\cite{gorur2010dirichlet} to cluster the dataset. DPGMM is an infinite mixture model with the Dirichlet Process as a prior distribution on the number of clusters, so it does not need predefined cluster number. 

In the experiments, we generate trajectories based on the scenarios in the Argoverse validation set. For each scenario, we generate six trajectories that are: 1) the most likely, 2) aggressive, 3) conservative, 4) turning (changing lane to) left, 5) turning (changing lane to) right, and 6) moving forward. We compare the results with trajectories generated by other representative and state-of-the-art trajectory generator/predictors including GCN-based~\cite{liang2020learning}, transformer-based~\cite{liu2021multimodal} and autoencoder-based works. All the results are in the same scenarios and of the same number of trajectories. We only count the clusters containing more data than the threshold ratio. The result in Table~\ref{tab:diversity} shows that \textbf{our model can generate more diverse and complex scenarios} based on past reference trajectories.  Our model significantly outperforms other methods, especially when the threshold is high, which means that our model can effectively augment rare behaviors and scenarios in the dataset (e.g., changing lane on the straight road), and balance their distribution. This augmentation can benefit the training and evaluation of prediction and planning modules.

\begin{table}[h]
\centering
\caption{Generation Diversity of Argoverse Validation Dataset}   

\label{tab:diversity}
\begin{tabular}{|c|c|c|c|}
\hline
\diagbox{Model}{Threshold}    & 0.05$\uparrow$ & 0.03$\uparrow$ & 0.01$\uparrow$    \\ \hline
GCN+Multi-head predictor\cite{liang2020learning}  & $6$  &$9$ & $29$  \\ \hline
mmTransformer\cite{liu2021multimodal}  & $2$  &$6$ & $33$ \\ \hline
Vanilla AAE  & $2$  &$5$ & $18$ \\ \hline
Ours  &  $10$ &$13$& $35$   \\ \hline

\end{tabular}
\end{table}

\begin{figure*}[htbp]
\centering

\subfigure[]{
\begin{minipage}[t]{0.2\linewidth}
\centering
\includegraphics[width=1.6in]{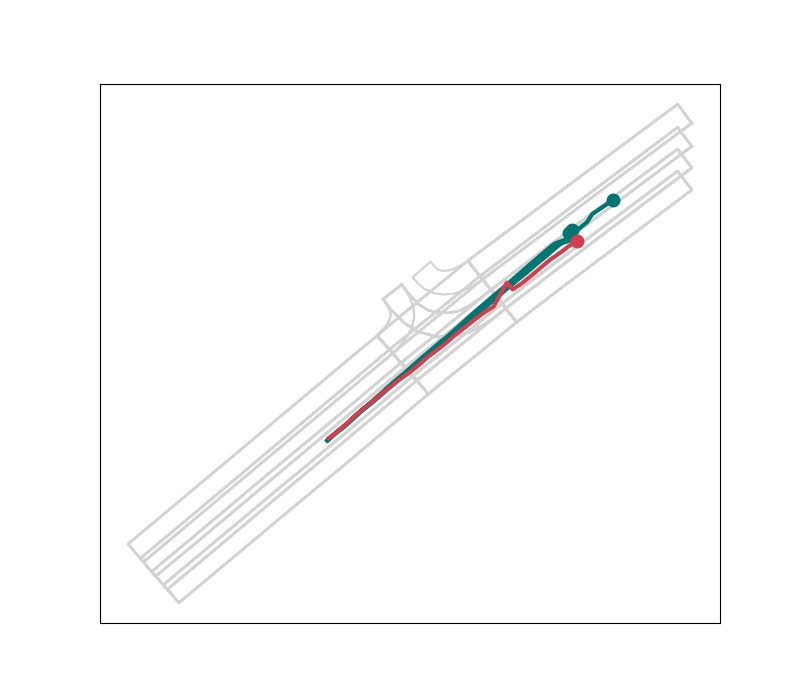}
\end{minipage}%
}%
\subfigure[]{
\begin{minipage}[t]{0.2\linewidth}
\centering
\includegraphics[width=1.6in]{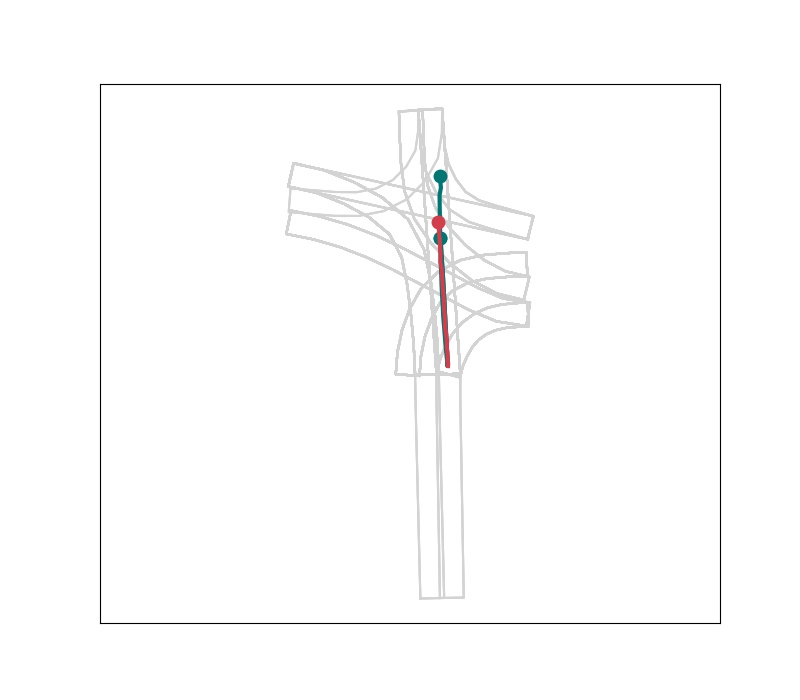}
\end{minipage}%
}%
\subfigure[]{
\begin{minipage}[t]{0.2\linewidth}
\centering
\includegraphics[width=1.6in]{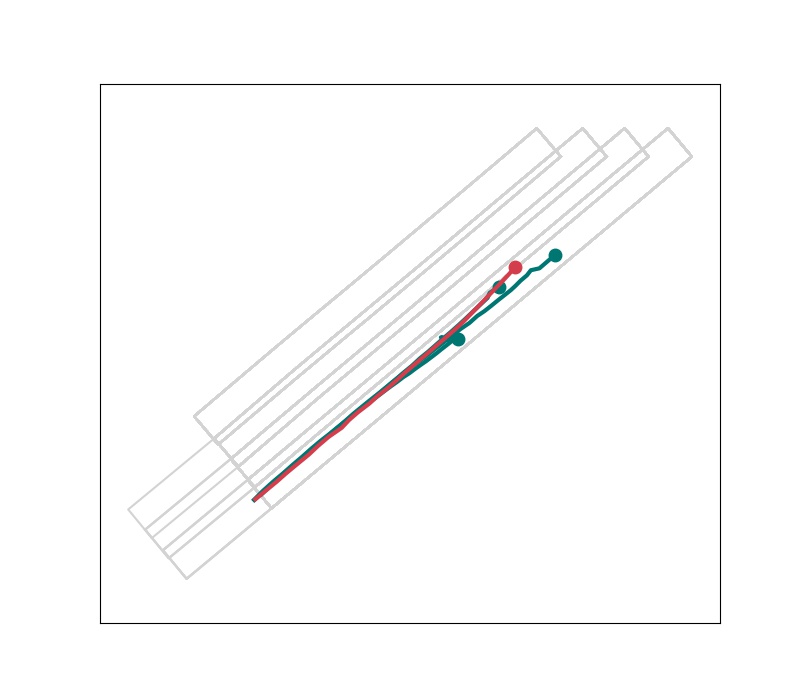}
\end{minipage}%
}%
\subfigure[]{
\begin{minipage}[t]{0.2\linewidth}
\centering
\includegraphics[width=1.6in]{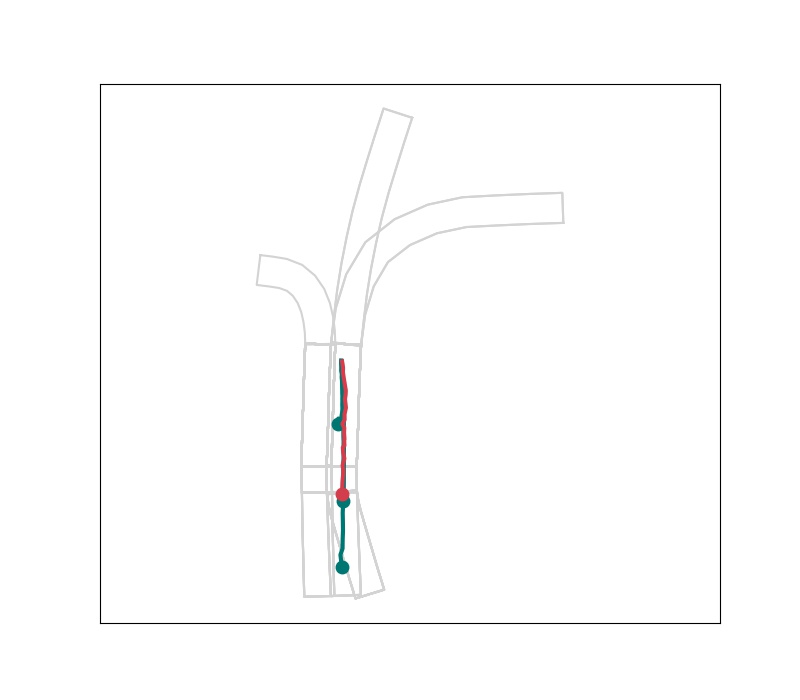}
\end{minipage}%
}%
\centering
\caption{ Trajectories generated with different levels of aggressiveness. The green trajectories are generated ones and the red trajectories are the references. }
\label{fig:agg}
\end{figure*}

\begin{figure*}[htbp]
\centering

\subfigure[]{
\begin{minipage}[t]{0.2\linewidth}
\centering
\includegraphics[width=1.6in]{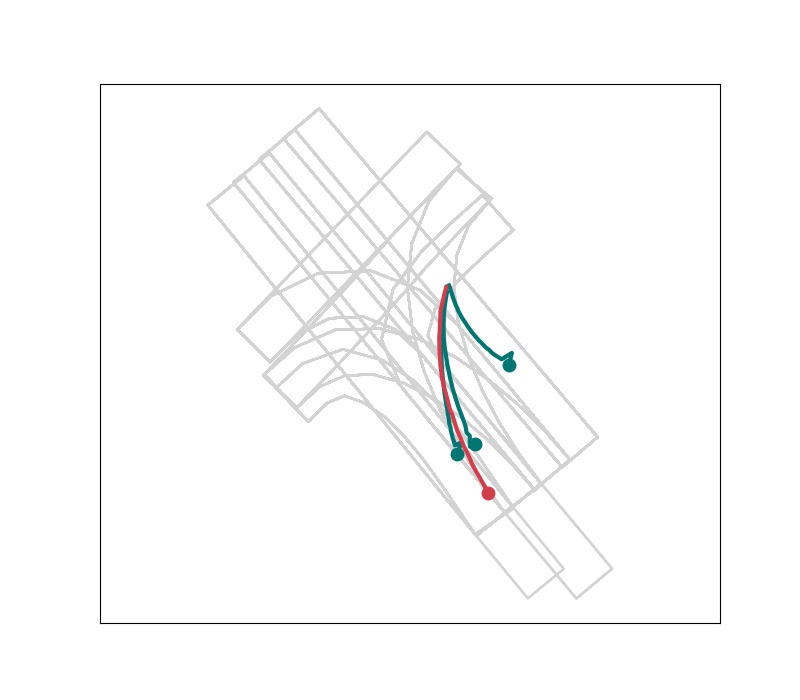}
\end{minipage}
}%
\subfigure[]{
\begin{minipage}[t]{0.2\linewidth}
\centering
\includegraphics[width=1.6in]{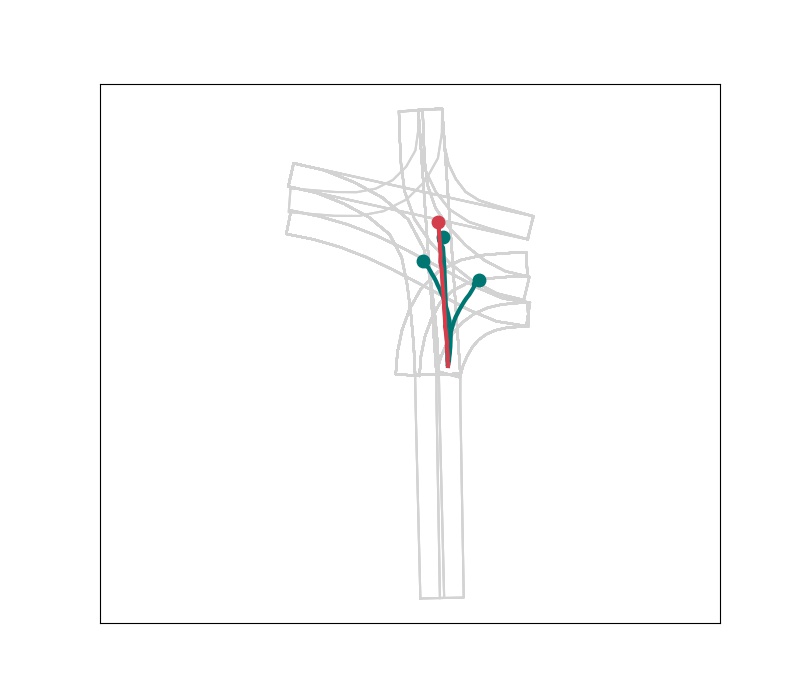}
\end{minipage}
}%
\subfigure[]{
\begin{minipage}[t]{0.2\linewidth}
\centering
\includegraphics[width=1.6in]{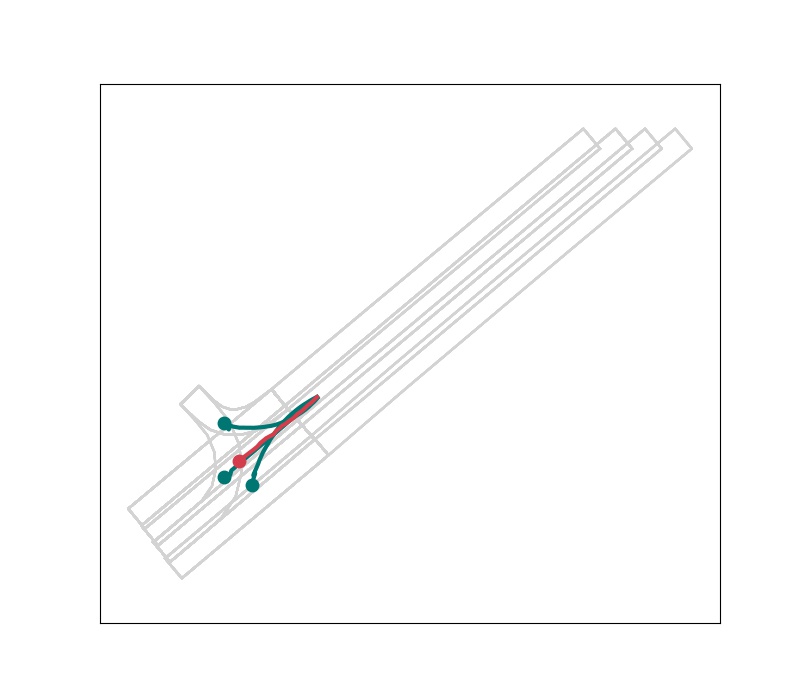}
\end{minipage}
}%
\subfigure[]{
\begin{minipage}[t]{0.2\linewidth}
\centering
\includegraphics[width=1.6in]{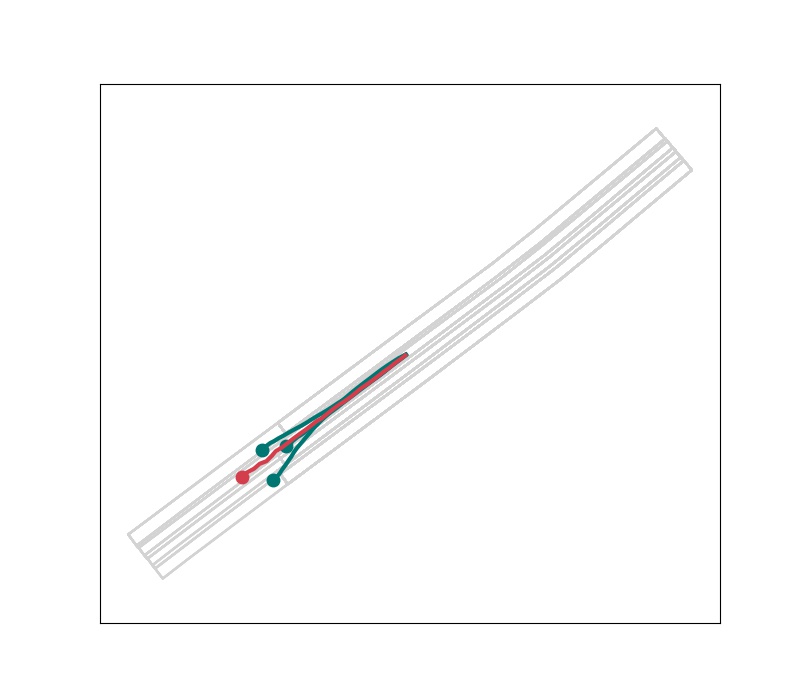}
\end{minipage}
}
\centering
\caption{Trajectories generated with different intentions. The green trajectories are generated ones and the red trajectories are the references. We generate 1) most-likely trajectory, 2) turning/changing lane to the left, 3) turning/changing lane to the right.}
\label{fig:intention}
\end{figure*}

Second, for the visualization, after inputting the past trajectories and road map, we adjust the aggressiveness (Fig.~\ref{fig:agg}) and intention (Fig.~\ref{fig:intention}), respectively. We can find that the behaviors are represented and disentangled in the latent space. The change of intention mainly leads to turning or lane changing according to the contexts. And the change of aggressiveness can be decoded to different accelerations in the longitudinal direction. 

Finally, to test the capability of generating safety-critical scenarios, we count the number of risky scenarios with different aggressiveness and intention settings. We define the risky scenario as situations where closest distance between two vehicles is less than 0.5 meter. We assume that the ego vehicle has an ideal planner that exactly follows the reference trajectory in the dataset while we can manually change other vehicles' behaviors by TAE. We first get the most-likely generation result with its aggressiveness ($agg$) and sweep the aggressiveness from most conservative to most aggressive. The result in Table~\ref{tab:safety-critical} shows that more aggressive behaviors will cause more risky situations on the road \emph{exponentially}. A conservative vehicle will be safer in general, although, being too conservative could also be unfavorable for safety, which matches our driving experience.  
We also switch the intention to the values representing more actively turning or lane changing behavior and observe an increase of the risky scenarios by $35.5\%$.

\begin{table}[h]
\centering
\caption{Safety critical scenarios changes }   

\label{tab:safety-critical}
\begin{tabular}{|c|c|c|c|c|c|}
\hline
$agg-3$ & $agg-2$ & $agg-1$ & $agg+0.5$ & $agg+1$ & $agg+1.5$  \\\hline
$-10.0\%$ &$-10.8\%$  &$-6.0\%$ & $+8.4\%$ &$+65.9\%$ &$+227.5\% $\\ \hline
\end{tabular}
\vspace{-6pt}
\end{table}

\subsection{Behavior-aware Trajectory Prediction}
Without manipulating in the latent space, we can directly obtain the behavior-aware motion predictor. We evaluate the accuracy of 1) behavior prediction in the latent space, and 2) most-likely trajectory in the final stage. The Table~\ref{tab:behavior} shows the performance of behavior prediction. The intention prediction can be regarded as a classification problem and the accuracy of our approach is 89.16\%. 
For the aggressiveness prediction, we use mean square error (MSE) to measure the accuracy. Our model can achieve an average MSE of 0.36 $s^2$, given the standard deviation of the aggressiveness over the dataset is 0.96.

\begin{table}[h]
\centering
\caption{Behavior Prediction Results }   
  
\label{tab:behavior}
\begin{threeparttable}  
\begin{tabular}{|c|c|c|c|}
\hline
 &  Accuracy/MSE    \\ \hline

Intention $\uparrow$  &  $89.16\%$  \\ \hline
Aggressiveness $\downarrow$ & 0.36 \\ \hline
\end{tabular}

\end{threeparttable}
\end{table}

To assess the average performance of trajectory prediction, we measure the average displacement error (ADE) between predicted and ground truth waypoints, and final displacement error (FDE) between last-predicted and ground truth waypoint.  Table~\ref{tab:pred} shows the results of our model and recent state-of-the-art works. Despite focusing more on long-tail events and diverse trajectory generation, our model achieves similar prediction performance in these average metrics and the results show that our model can generate natural and realistic trajectories based on a small latent space.   
\begin{table}[h]
\centering
\caption{Prediction Results }   

\label{tab:pred}
\begin{tabular}{|c|c|c|c|c}
\hline
\diagbox{Model}{Metrics}   & ADE$\downarrow$ & FDE$\downarrow$    \\ \hline
Argoverse Baseline (NN)\cite{chang2019argoverse} & $3.45$ &$7.88$  \\ \hline
Jean\cite{mercat2020multi}  & $1.74$  &$4.24$   \\ \hline
TNT\cite{zhao2020tnt}  & $1.77$  &$3.91$  \\ \hline
LaneGCN\cite{liang2020learning}  & $1.71$  &$3.78$  \\ \hline
WIMP\cite{khandelwal2020if} & $1.82$  &$4.03$ \\ \hline
TPCN\cite{ye2021tpcn} & $1.66$  &$3.69$  \\ \hline
Ours  &  $1.73$ &$3.83$  \\ \hline

\end{tabular}
\end{table}

\subsection{Discussions}
By explicitly modeling the behavior-level vehicle intention and aggressiveness in the latent space, our framework can provide more diverse and controllable trajectory generation as well as good prediction performance in a unified architecture. And we believe that the semi-supervised latent space modeling can be extended to more behaviors. 

During experiments, we have a few observations that can be further explored.
 In latent space modeling, we find that there exist distribution imbalances in some attributes like intentions (more than 60\% scenarios are moving forward), and it reveals a promising direction of improving the trajectory modeling. We demonstrate that our model can produce realistic trajectories with a smaller latent space compared to other works. However, there still exists an information loss when encoding features to the lower dimensional latent space and this is part of the reason why works such as TPCN~\cite{ye2021tpcn} have better prediction displacement error than ours. 
In future work, we plan to utilize our framework to further model more driving behaviors, test other context extractors, and add safety constraints to the current framework.
